\documentclass[10pt,twocolumn,letterpaper]{article}

\usepackage{cvpr}              

\usepackage[accsupp]{axessibility}
\usepackage[dvipsnames]{xcolor}

\usepackage{graphicx}
\usepackage{amsmath}
\usepackage{amssymb}
\usepackage{booktabs}
\usepackage{bbding}
\usepackage{pifont}
\usepackage{wasysym}
\usepackage{amssymb}
\usepackage{soul}

\definecolor{cvprblue}{rgb}{0.21,0.49,0.74}
\usepackage[pagebackref,breaklinks,colorlinks,citecolor=cvprblue]{hyperref}
\usepackage{amsmath}
\usepackage{xcolor,colortbl}
\def\paperID{6326} 
\def\confName{CVPR}
\def\confYear{2025}

\author{Rolandos Alexandros Potamias$^1$, Jinglei Zhang$^2$, \\Jiankang Deng$^1$, Stefanos Zafeiriou$^1$ \\
$^1$Imperial College London, $^2$Shanghai Jiao Tong University\\
}

\begin{document}
\title{WiLoR: End-to-end 3D Hand Localization and Reconstruction in-the-wild }
\twocolumn[{
\renewcommand\twocolumn[1][]{#1}%
\maketitle
\begin{center}
    \centering
    \captionsetup{type=figure}
    \includegraphics[width=0.95\textwidth]{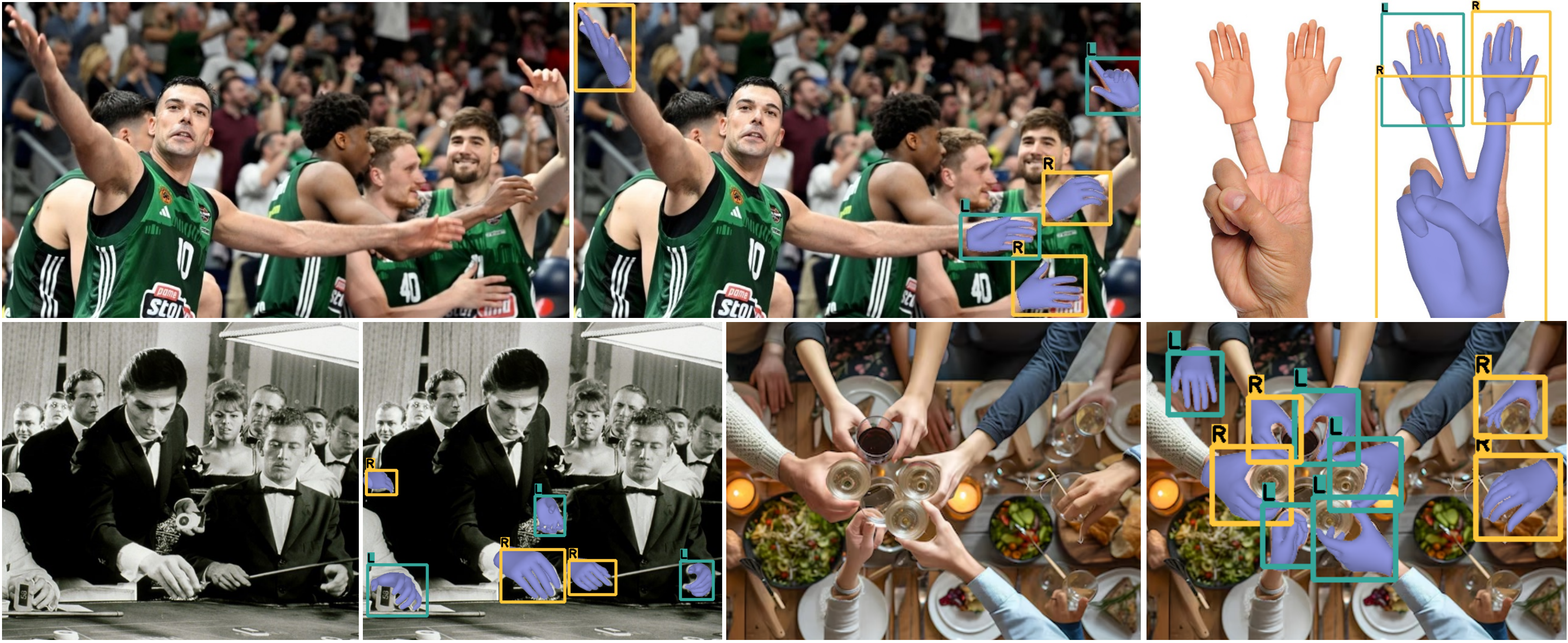}
    \captionof{figure}{We propose \textbf{WiLoR}, a full-stack in-the-\textbf{Wi}ld \textbf{Lo}calization and 3D hand \textbf{R}econstruction method. WiLoR first localizes and defines the handedness of the detected hands which are then lifted to 3D using a transformer-based hand pose estimation module. To aid high-fidelity reconstructions and facilitate image-alignment, we introduce a refinement module that extracts localized features to correct misaligned poses. WiLoR achieves state-of-the-art performance under different benchmark datasets while boosting the temporal coherence of image-based 3D hand pose estimation methods.}
\end{center}}]
\begin{abstract}
In recent years, 3D hand pose estimation methods have garnered significant attention due to their extensive applications in human-computer interaction, virtual reality, and robotics. In contrast, there has been a notable gap in hand detection pipelines, posing significant challenges in constructing effective real-world multi-hand reconstruction systems. In this work, we present a data-driven pipeline for efficient multi-hand reconstruction in the wild. The proposed pipeline is composed of two components: a real-time fully convolutional hand localization and a high-fidelity transformer-based 3D hand reconstruction model. To tackle the limitations of previous methods and build a robust and stable detection network, we introduce a large-scale dataset with over than 2M in-the-wild hand images with diverse lighting, illumination, and occlusion conditions. Our approach outperforms previous methods in both efficiency and accuracy on popular 2D and 3D benchmarks. Finally, we showcase the effectiveness of our pipeline to achieve smooth 3D hand tracking from monocular videos, without utilizing any temporal components. 
Code, models, and dataset are available on our \href{https://rolpotamias.github.io/WiLoR}{project page}.

\end{abstract}    
\section{Introduction}
\label{sec:intro}
Hand detection and reconstruction has been a long-studied problem due to its numerous applications, ranging from virtual reality~\cite{grauman2022ego4d} to sign language~\cite{Baltatzis_2024_CVPR,zuo2024signs,hu2021signbert} and human behaviour recognition~\cite{qi2024computer}. Given the large variations in hand appearance and articulation~\cite{Potamias_2023_CVPR} along with heavy occlusions and motion blur~\cite{hampali2020honnotate,hampali2022keypointtransformer} that are usually present in hand interactions, the task of hand pose estimation is considerably challenging. Over the years, several methods have been proposed to tackle 3D hand pose estimation~\cite{moon2020i2l,choi2020pose2mesh,lin2021mesh,pavlakos2024reconstructing}. However, despite producing credible results, these methods primarily focus on images containing a fixed number of hands and hence cannot generalize to in-the-wild images.

In the closely related fields of 3D human body and face reconstruction, state-of-the-art methods~\cite{cao2017realtime,alphapose,guo2020towards,goel2023humans} employ bottom-up pipelines founded on top of high-performance detection models that initially localize human body and face within the image, enabling their generalization to in-the-wild images. 
Despite the numerous methods that have been proposed to solve the task of human body and face detection, there has been a notable lack in real-time hand detection methods.
The importance of hand detectors is further emphasized considering that current 3D hand pose estimation frameworks operate on tight crops around the hand regions~\cite{MobRecon,simpleHand}.
Consequently, hand detectors are essential for the generalisation of such methods in in-the-wild scenarios.
Popular hand detection and localisation methods~\cite{zhang2020mediapipe,cao2017realtime} fail significantly to detect multiple hands and challenging poses, while more recent methods~\cite{narasimhaswamy2019contextual,narasimhaswamy2020detecting,li2022exploring} albeit producing reasonable results struggle to operate in real-time. 
Motivated by the lack of accurate hand detection frameworks, we propose a robust single-state anchor-free detector that can operate in over than 100 frames-per-second (fps). As we experimentally show, robust detections can enforce more stable 4D reconstructions and overcome jittering artifacts which is currently one of the main limitations of 3D frame-based pose estimation methods.

In contrast to the relatively unexplored hand detection and localization problem, 3D hand pose estimation has received significantly more attention. 
Initial 3D pose reconstruction methods have focused on traditional convolution-based backbones to process and extract image features~\cite{boukhayma20193d,kulon2020weakly,moon2020i2l,choi2020pose2mesh}. 
Following the success of transformers and their ability to consume large amounts of data~\cite{vaswani2017attention,devlin2018bert,lee2021vitgan,rombach2021diffusion,tarasiou2024locally,zhang2025hawor}, several methods have paved the way of utilising transformer architectures scaling up the 3D human body and hand recovery~\cite{lin2021end,lin2021mesh,pavlakos2024reconstructing}.
Recently, Pavlakos \etal~\cite{pavlakos2024reconstructing} showcased the effectiveness of vision transformers (ViT) using a simple yet powerful framework trained on a large-scale dataset. 
The key to the success of this method lies in the scale of its architecture, composed of more than 0.5 billion parameters, enabling effective consumption of large amount of data. 
However, as shown in the literature~\cite{stacked_hourglass,sun2019deep,OSX,zhang2023pymaf}, regressing the hand parameters from a single image results in poor alignment and incorrect poses. 
Currently, methods that aim to achieve better image alignment rely on sub-optimal solutions, such as intermediate heatmap representations~\cite{PointHMR,MobRecon,simpleHand}.  
To tackle this, we propose a high-fidelity 3D pose estimation method that decomposes 3D hand reconstruction into two stages. In particular, the decoder first predicts a rough hand estimation that is used to extract multi-scale image-aligned features from our refinement module. By leveraging the rough hand estimation, we can extract meaningful spatial features that lead to better image alignment and state-of-the-art performance on FreiHand~\cite{freihand} and HO3D~\cite{hampali2020honnotate} benchmark datasets. Additionally, 
in contrast to vertex regression methods~\cite{kulon2020weakly,lin2021end,lin2021mesh} that directly regress 3D vertices, our method predicts MANO parameters~\cite{mano}, ensuring both explainable and plausible hand poses. 



In this paper, we propose a high fidelity full stack method that can reconstruct 3D hands in real-time. Specifically: 
\begin{itemize}
    \item Based on the limitations of current hand detection benchmark datasets, we collect a large-scale dataset of in-the-wild images that contain multiple hands and introduce a challenging benchmark for hand detection. We make the dataset, along with the corresponding 2D and 3D annotations, publicly available. 
    
    \item We propose a real-time hand detection method trained on the collected large-scale dataset that outperforms previous hand detection methods by a large margin in both accuracy and efficiency. 

    \item We propose a transformer-based method that facilitates high fidelity 3D reconstructions and tackles the architectural limitations of previous methods using a novel refinement module. The proposed method, apart from highly efficient, achieves state-of-the-art performance in both Freihand and HO3D benchmark datasets. 
    
\end{itemize}




\section{Related Work}
\label{sec:related}

\textbf{Hand Detection and Tracking.}
Object detection has been extensively studied in the literature achieving remarkable advancements~\cite{liu2016ssd,retinanet,redmon2018yolov3} and setting the foundations for human body~\cite{sun2011articulated,pishchulin2012articulated,gkioxari2014using,papandreou2017towards,alphapose,xu2022vitpose} and face detection~\cite{retinaface,zhu2020tinaface} pipelines. 
In contrast, despite two decades of research efforts~\cite{rehg1994digiteyes}, hand detection has not yet achieved comparable breakthroughs. Initial approaches used controlled conditions and depth cameras~\cite{wu2000adaptive,zhu2000segmenting,stenger2006model,wang2009real,sridhar2013interactive} to detect and track human hands. Several efforts have been made to boost hand detection under different skin tones and backgrounds using multi-stage frameworks~\cite{mittal2011hand,pisharady2013attention}, however, they fail to generalize in challenging environments. Following the success in object detection, several methods have adopted fully convolutional architectures for hand detection~\cite{hoang2016multiple,roy2017deep,deng2017joint,zhang2020mediapipe,shan2020understanding}. Simon \etal~\cite{simon2017hand} introduced a multi-view bootstrapping procedure to annotate in-the-wild data and train a real-time convolutional detector network. 
Recently, Narasimhaswamy \etal~\cite{narasimhaswamy2019contextual} proposed an extension of MaskRCNN~\cite{he2017mask} network to detect in-the-wild hands and identifying their corresponding contact points~\cite{narasimhaswamy2020detecting} and body associations~\cite{narasimhaswamy2022whose}. Nevertheless, despite the extensive efforts in the literature, most methods rely on slow backbones and struggle with challenging images. The primary issue is the lack of large-scale training data featuring multiple levels of occlusions and motion blur from in-the-wild scenes. To tackle such limitation, we propose a lightweight hand detector that is 45 $\times$ faster compared to previous state-of-the-art detectors, trained on 2M in-the-wild images with diverse environments and occlusion. 

\noindent\textbf{3D Hand Pose Estimation.} 
Similar to hand detection, initial approaches for hand pose estimation relied on depth cameras~\cite{oikonomidis2011efficient,tagliasacchi2015robust,ge2016robust} to reconstruct 3D hands. Boukhayma \etal \cite{boukhayma20193d} introduced the first fully learnable pipeline that directly regresses the parameters of the MANO hand model~\cite{mano} from RGB images. In a similar manner, several follow-up works used heatmaps~\cite{zhang2019end} and iterative refinement~\cite{baek2019pushing} to enforce 2D alignment. 
Kulon \etal~\cite{kulon2019single,kulon2020weakly} introduced an alternative regression method that directly regresses 3D vertices using spiral graph neural networks \cite{bouritsas2019neural,potamias2022graphwalks}, which significantly outperformed previous methods. Various approaches have been proposed to improve task-specific challenges of 3D pose estimation, including robustness to occlusions~\cite{HandOccNet} and motion blur~\cite{BlurHand} and reducing inference speed~\cite{MobRecon,simpleHand}. Recently, Pavlakos \etal~\cite{pavlakos2024reconstructing} highlighted the importance of scaling up both the training data and the capacity of the model. Specifically, building on the success of the Vision Transformer (ViT) backbones for body pose estimation~\cite{OSX,4dhumans,smpler-x}, they demonstrated that using a simple yet effective large-scale transformer architecture can achieve state-of-the-art performance when trained on a diverse collection of datasets. However, directly regressing MANO parameters from the image in one go may introduce misalignments and incorrect poses. To tackle this, we propose a novel refinement layer that deforms hand pose using mesh-aligned multi-scale features.


\section{WHIM Dataset}
\label{sec:dataset}
A vital cause behind the lack of high-fidelity hand detection systems lies in the limited amount of in-the-wild datasets with multiple hand annotations. To build a robust hand detection framework, we collected a large-scale dataset with \textbf{m}illions of \textbf{i}n-the-\textbf{w}ild \textbf{h}ands (\textbf{WHIM}) with diverse poses, illuminations, occlusions, and skin tones. 

To collect the proposed dataset, we devised a pipeline to automatically annotate YouTube videos from diverse and challenging in-the-wild scenarios. In particular, we selected more than 1,400 YouTube videos containing hand activities including sign language, cooking, everyday activities, sports, and games with ego- and exo-centric viewpoints, motion blur, different hand scales, and interactions. To accurately detect and annotate the hands on each frame we used a combination of ensemble networks. First, we used VitPose~\cite{xu2022vitpose} and AlphaPose~\cite{alphapose} to detect all humans in the frame and selected the bounding boxes with confidence bigger than 0.65. We then cropped the bounding boxes and fed them to an ensemble hand detection pipeline that consists of MediaPipe~\cite{zhang2020mediapipe}, OpenPose~\cite{cao2017realtime} and ContactHands~\cite{narasimhaswamy2019contextual} models. To localize the hand, we used a weighted average between the bounding box positions $\mathbf{b}_i$ of the three detectors $d_i$, scaled from their corresponding confidence $P(\mathbf{b}_i|d_i)$: 
\begin{equation}
    \hat{y} = \frac{\sum_i P(\mathbf{b}_i|d_i) \mathbf{b}_i }{\sum_i  P(\mathbf{b}_i|d_i)}
\end{equation}
where ${\mathbf{b}_i}$ denotes the estimated hand bounding box. 

In addition to the bounding box, we used the estimated 2D landmarks~\cite{zhang2020mediapipe,cao2017realtime}, to fit a 3D parametric hand model $\mathcal{M}$~\cite{mano}. More specifically, we optimized shape $\mathbf{\beta}$ and pose $\mathbf{\theta}$ parameters to minimize the re-projection loss $\mathcal{L}_{proj}$ between the regressed ${\mathbf{J}_\mathcal{M}}$ and the estimated landmarks $\mathbf{\hat{J}_s}$: 
\begin{equation}
    \mathcal{L}_{proj} = ||\mathbf{J}_\mathcal{M} - \pi(\mathbf{\hat{J}_s}, K)||_1,
\end{equation}
where $\pi(\cdot)$ denotes the weak perspective projection transform and $K$ the estimated intrinsic camera matrix. 

Given the degrees of freedom of the human hand, optimizing the hand model using joint terms usually results in unnatural poses. To tackle the ambiguities during the optimization process, we followed~\cite{spurr2020weakly} and included bio-mechanical losses to constrain the optimization. In particular, apart from the re-projection error, we enhanced the fitting process using loss functions that constrain the bone lengths and the angle rotations to feasible ranges, as defined in~\cite{spurr2020weakly}: 
\begin{equation}
    \mathcal{L}_{BMC} = \mathcal{L}_{BL} + \mathcal{L}_{A}
\end{equation}
where $\mathcal{L}_{BL}$ and $\mathcal{L}_{A}$ denote bone length and joint angle loss terms, respectively. For additional details of the bio-mechanical constraints we refer the reader to~\cite{spurr2020weakly}. 

\begin{figure}[!h]
    \centering
\includegraphics[width=\linewidth]{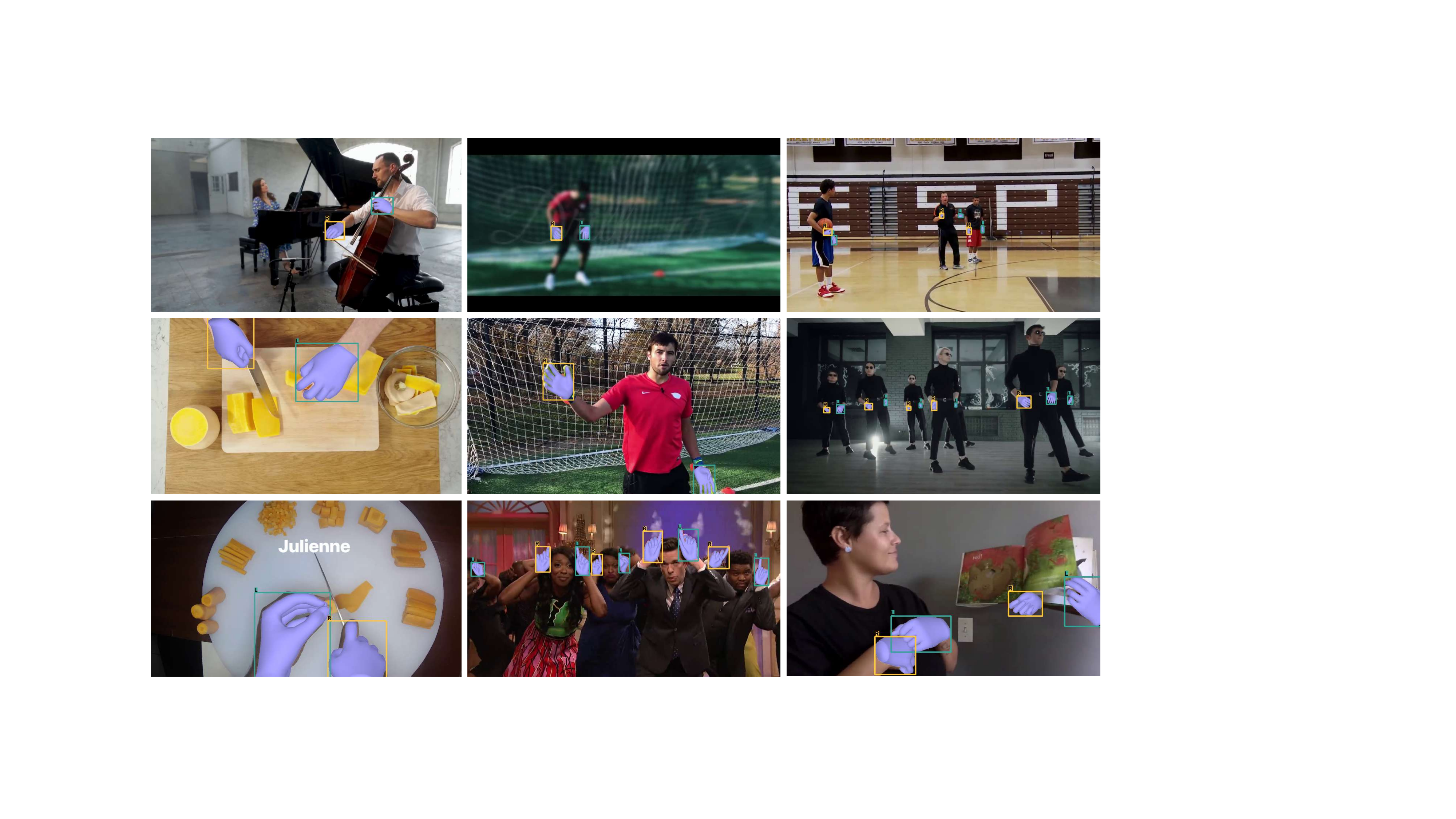}
    \captionof{figure}{Example of the proposed WHIM in-the-wild dataset.
    \label{fig:dataset_example}}
\end{figure}

Finally, given that the bio-mechanical prior acts mainly on the joint space, we followed~\cite{Baltatzis_2024_CVPR} and trained a PCA model on ARCTIC dataset~\cite{fan2023arctic} acting as a 3D prior, modeling the distribution of feasible hand poses. 
We  formulated the prior loss as the reconstruction error of the 3D mesh $\mathbf{X}$ projected and reconstructed from the PCA space $\mathbf{U}$, as: 
\begin{equation}
    \mathcal{L}_{prior} = || \mathbf{X} -  [(\mathbf{X} - \boldsymbol{\mu} )\mathbf{U} ^ T] \mathbf{U} + \boldsymbol{\mu} ||_2,
\end{equation}
where $\mathbf{U} \in \mathbb{R}^{d \times N\cdot3}$ denotes the eigenvector basis of $d$ components and $\boldsymbol{\mu}$ the mean mesh. \cref{fig:dataset_example} includes several examples of the proposed WHIM dataset.

\section{Method}
\label{sec:method}
\subsection{Hand Detection and Localization}
\begin{figure}[!ht]
    \centering
    \includegraphics[width=\linewidth]{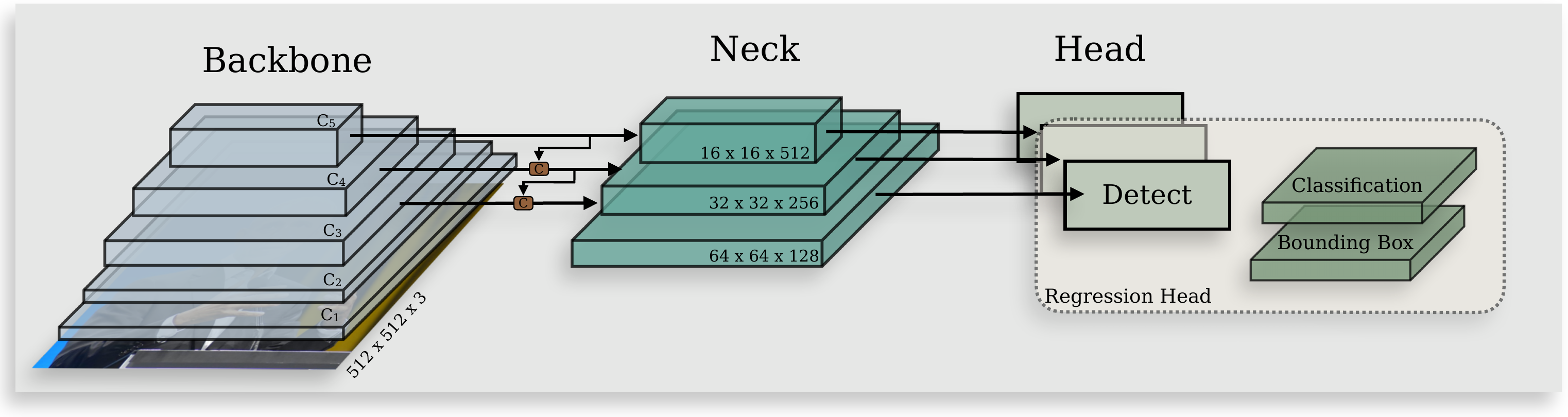}
    \captionof{figure}{
    \textbf{Detection overview}: The proposed fully convolutional one-stage hand detection method receives an image and extracts multi-resolution feature maps that are then processed by the Path Aggregation Network (PANet). The corresponding features are then fed to three detection heads that predict the hand side, bounding box, and hand joints at different resolutions. We train the network with a multi-task loss for each anchor. 
    \label{fig:detection}}
\end{figure}

Over the past years, fully convolutional networks (FCNs) have shown remarkable efficiency in human detection~\cite{wang2020deep} and object detection~\cite{redmon2016you}. Building on their success, we employ an FCN architecture to achieve both accurate and real-time hand localization.
Similar to object detection frameworks, given an image $\mathbf{I} \in \mathbb{R}^{H \times W \times 3}$ our goal is to detect the bounding boxes $\mathbf{B}=\{\mathbf{b}_j \in \mathbb{R} ^ 4 : 0 \le j \le n\} $ of the $n$ hands present in the image along with their hand side label $\mathbf{y}_j$.
We follow the commonly used one-stage backbone-neck-head formulation and we built upon the powerful and efficient DarkNet backbone~\cite{redmon2018yolov3}.  
We extract the last three feature maps $\{C3, C4, C5\}$ of the backbone to generate a multi-scale feature pyramid in the neck module. 
To enable our model to effectively capture multi-scale features using both top-down and bottom-up pathways across different feature maps, we utilized Path Aggregation Network (PANet)~\cite{liu2018path}, an extension of Feature Pyramid Network~\cite{lin2017feature} that facilitates fine-grained information flow using a bottom-up path augmentation. 
Finally, we use three detection heads to predict the bounding boxes $\mathbf{b}_j$ and hand side labels $\mathbf{y}_j$ at different anchor resolutions. 
Following~\cite{duan2019centernet}, we adopt an anchor-free design to enhance the flexibility of our localization method and directly predict bounding box coordinates without relying on predefined anchor boxes.
An overview of the proposed detection network is visualized in \cref{fig:detection}. 
Similar to~\cite{retinaface}, we observed that joint keypoint supervision significantly improved the performance and the robustness of the detector. The full training objective can be defined as: 
\begin{equation}
    \mathcal{L} = \lambda_{0}\mathcal{L}_{BCE} + \lambda_{1}\mathcal{L}_{DFL} + \lambda_{2}\mathcal{L}_{CIoU} + \lambda_{3}\mathcal{L}_{kpts} 
\end{equation}
where $\mathcal{L}_{BCE}$ is the binary cross entropy loss between the predicted and the ground truth box labels, $\mathcal{L}_{DFL}$ denotes the distributional focal loss~\cite{li2020generalized} which measures the difference between the predicted and the ground truth bounding box distributions, $\mathcal{L}_{CIoU}$ measures the discrepancy between the predicted and the ground truth bounding box~\cite{zheng2021enhancing},  $\mathcal{L}_{kpts}$ denotes an $L2$ loss on the  and $\lambda_{0},\lambda_{1},\lambda_{2},\lambda_{3}$ are weights that balance the losses. 

\begin{figure*}[!h]
    \centering
    \includegraphics[width=0.93\textwidth]{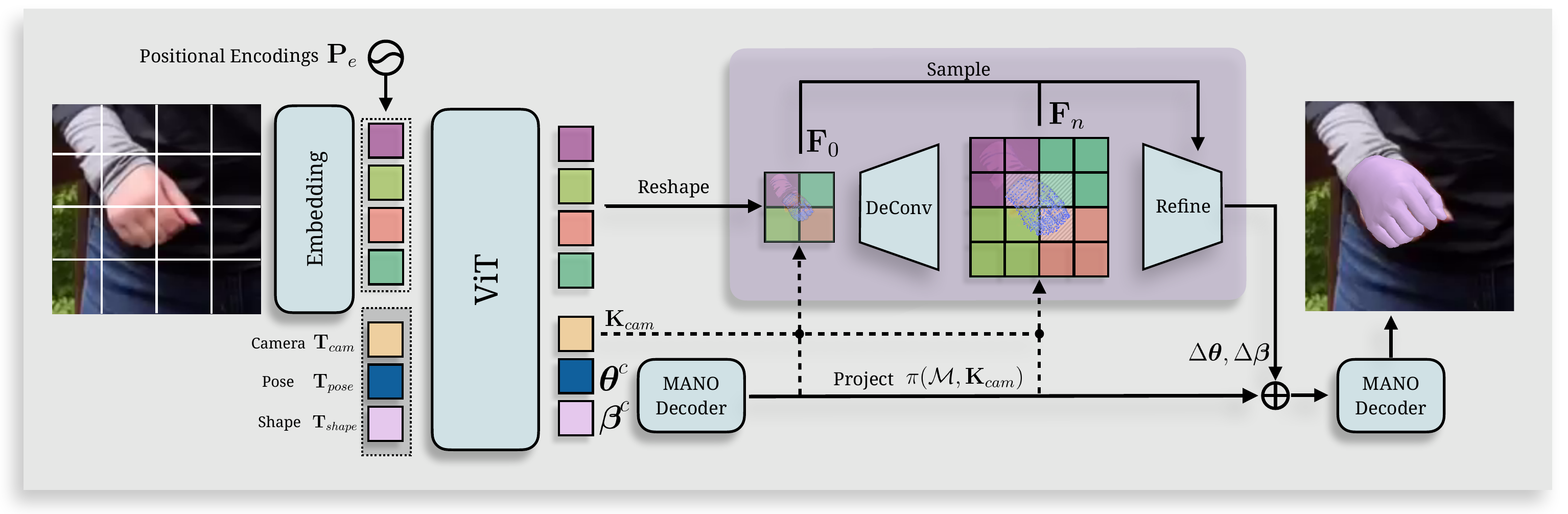}
    \captionof{figure}{
    \textbf{Overview of the proposed 3D hand pose estimation method}: Given an image $\mathbf{I}_h$ represented as a series of feature tokens $\mathbf{T}_{img}$ along with a set of learnable camera $\mathbf{T}_{cam}$, pose $\mathbf{T}_{pose}$ and shape $\mathbf{T}_{shape}$ tokens, we initially predict a rough estimation of the MANO~\cite{mano}  and camera $\mathbf{K}_{cam}$ parameters using a ViT backbone (\textcolor{cyan}{light blue}). The updated image tokens are then reshaped and upsampled through a series of deconvolutional layers to form a set of multi-resolution feature maps $\{\mathbf{F}_{0},...,\mathbf{F}_{0}\}$. We then project the estimated 3D hand to the generated feature maps and sample image-aligned multi-scale features through a novel refinement module \textcolor{violet}{(purple)}. The sampled features are used to predict pose and shape residuals $\Delta\theta, \Delta\beta$ that refine the coarse hand estimation. Using this coarse-to-fine pose estimation strategy we facilitate image alignment and achieve better reconstruction performance. } 
    \label{fig:method}
\end{figure*}

\subsection{Hand Reconstruction}
Given an image $\mathbf{I_{h}} \in \mathbb{R}^{H \times W \times 3}$ that contains a human hand, tightly cropped around the hand detectors bounding box, the proposed 3D hand reconstruction method estimates the corresponding 
hand pose $\theta \in \mathbb{R}^{48}$ and shape $\beta \in \mathbb{R}^{10}$ MANO~\cite{mano} parameters along with the camera parameters $\mathbf{K}_{cam} = \{\mathbf{t}_{cam}, \mathbf{s}_{cam}\} $ to obtain a 3D hand.  

To build a powerful 3D pose estimation network that can scale on large amounts of data, we follow~\cite{OSX,smpler-x,pavlakos2024reconstructing}, and built our backbone using a pre-trained ViT encoder~\cite{vaswani2017attention,xu2022vitpose}. The image $\mathbf{I_{h}}$ is first split into $M$-size patches $\mathbf{P} \in \mathbb{R}^{\frac{HW}{M^2}\times (M^2 \times 3)}$ and then embedded to high dimensional tokens $\mathbf{T}_{img} \in \mathbb{R}^{\frac{HW}{M^2}\times C}$. To uniquely encode their spatial location, positional embeddings $\mathbf{P}_e$ are added to the image tokens $\mathbf{T}_{img}$~\cite{vaswani2017attention}. In addition to the image tokens, we explicitly model hand pose, shape, and camera parameters with three distinct tokens $\mathbf{T}_{pose}, \mathbf{T}_{shape}, \mathbf{T}_{cam}$. We then feed the concatenated tokens to the ViT transformer encoder to obtain a set of updated feature tokens $\mathbf{T'}_{img}, \mathbf{T'}_{pose}, \mathbf{T'}_{shape}, \mathbf{T'}_{cam}$. 
Using a set of MLP layers we regress a rough estimation of pose $\theta^{c}$ and shape $\beta^{c}$ parameters of the MANO model, which will serve as a prior for the refinement network. Similarly, we regress the camera translation and scale parameters $\mathbf{K}_{cam} = \{\mathbf{t}_{cam}, \mathbf{s}_{cam}\}$ from the camera token features. 

\noindent\textbf{Multi-Scale Pose Refinement Module.}
In order to get better image alignment and more accurate hand pose, we introduce a fully differentiable refinement module that predicts pose and shape residuals of the rough hand estimation. 
To achieve this, we utilize image features extracted from the ViT backbone as 2D feature cues within our refinement module. 
In particular, we reshape image feature tokens $\mathbf{T'}_{img}$ to form a low resolution feature map $\mathbf{F}_{0} \in \mathbb{R}^{\frac{H}{M}\times\frac{W}{M}\times C}$ and project the rough hand estimation $\mathcal{M}_{l}$ to feature map using the estimated $\mathbf{t}_{cam}, \mathbf{s}_{cam}$ camera parameters. 
Then, using bilinear interpolation we sample from $\mathbf{F}_{0}$ a feature vector $\mathbf{f}^\mathbf{v}_{0}$ for each projected vertex $\mathbf{v}$: 
\begin{equation}
    \mathbf{f}^\mathbf{v}_{0} = \pi(\mathbf{v}, \mathbf{K}_{cam}) 
\end{equation}
where $\pi(\cdot)$ denotes the weak perspective projection. 

Note that we project the whole hand mesh $\mathcal{M}_{l}$ to the feature map, instead of just the hand joints, as we aim to acquire better shape and pose image alignment. 
The image-aligned vertex features are then aggregated to form a global feature vector that is used to regress pose and shape residuals: 
\begin{equation}
\begin{aligned}
    \Delta\beta &= MLP_{\beta}(\square_{\mathbf{v} \in \mathcal{M}_{l}} \mathbf{f}^{\mathbf{v}}_{0}) \\  
    \Delta\theta&= MLP_{\theta}(\square_{\mathbf{v} \in \mathcal{M}_{l}} \mathbf{f}^{\mathbf{v}}_{0})
\end{aligned}
\end{equation}
where $\square$ denotes the aggregation function, \eg, mean, max, sum. 

Given that the initial feature map is very low-dimensional, we use a set of deconvolutional layers to upsample $\mathbf{F}_{0}$ to multiple higher resolution feature maps ${\mathbf{F}_{0},\mathbf{F}_{1},..., \mathbf{F}_{n}}$ that will serve as multi-scale features for the proposed refinement module. Intuitively, low-dimensional feature maps will provide global and structural residuals of the hand shape while more high-resolution features provide finer details of the hand pose.  

\noindent\textbf{Loss function.} The proposed model is trained with supervision for 3D vertices $\hat{\mathbf{V}}_{3D}$, 2D joints $\hat{\mathbf{J}}_{2D}$ as well as MANO parameters $\hat{\theta},\hat{\beta}$, when available. Additionally, following~\cite{kanazawa2018end,pavlakos2024reconstructing}, we utilize a discriminator network $D$ to enforce plausible hand poses and shapes and penalize irregular articulations. The full loss function can be defined as: 
\begin{equation}
\begin{aligned}
   &\mathcal{L} = \mathcal{L}_{3D} + \mathcal{L}_{2D}  + \mathcal{L}_{mano} + \mathcal{L}_{adv}, \\
    &\mathcal{L}_{3D} = ||\mathbf{V}_{3D} -\hat{\mathbf{V}}_{3D}||_1, \\
    &\mathcal{L}_{2D} = ||\pi(\mathbf{J}_{3D}, \mathbf{K}_{cam})  -\hat{\mathbf{J}}_{2D}||_1 \\
    &\mathcal{L}_{mano} = ||\theta - \hat{\theta}||_2^2 + ||\beta - \hat{\beta}||_2^2, \\
    &\mathcal{L}_\texttt{adv} = ||D(\theta,\beta) - 1||_2. \\
\end{aligned}
\end{equation}

\definecolor{tabfirst}{rgb}{1, 0.7, 0.7} 
\definecolor{tabsecond}{rgb}{1, 0.85, 0.7} 
\definecolor{tabthird}{rgb}{1, 1, 0.7} 

\section{Experiments}
\label{sec:experiments}
In this section we first evaluate the proposed hand detection network using established benchmarks to assess its performance. Next, we conduct an extensive qualitative and quantitative analysis of the proposed 3D hand pose estimation method. Finally, we demonstrate the critical role of precise hand localization in the accuracy of 4D hand reconstruction.

\subsection{Evaluation of Hand Detection and Localization}
\noindent{\textbf{Training.}} We train the proposed hand detection network using the curated WHIM dataset that consists of over 2M in-the-wild images of multiple hands and scales. To further boost the generalization and robustness of our network, we follow several data augmentations during training. Particularly, we introduce random rotations in the range of $[-60^{\circ}, 60^{\circ}]$ and translations in the range of $[-0.1, 0.1]$ along with random masking and cropping of the image. Additionally, in each training batch, we follow mosaic and mixup augmentation, which significantly affects the robustness to diverse hand scales. 

\noindent{\textbf{Evaluation.}} To compare our network, we employ popular baselines such as OpenPose~\cite{cao2017realtime} and Mediapipe~\cite{zhang2020mediapipe}, which are widely used across the community~\cite{qian2020html,Potamias_2023_CVPR}, along with more recent hand detection pipelines such as ContactHands~\cite{narasimhaswamy2019contextual} and ViTDet~\cite{li2022exploring}. All methods are evaluated under three criteria: i) the inference speed in terms of frames per second (FPS), ii) the detection performance in terms of average precision (AP) at IoU = 0.5 and mean AP at different IoU=0.5:0.05:0.95 thresholds and iii) the model size measured in Mb. 
An optimal hand detection system should be lightweight to ensure compatibility with mobile devices, operate in real-time to avoid impacting the runtime of a 3D pose estimation pipeline while achieving precise detections.
\begin{figure}[!h]
    \centering
\includegraphics[width=\linewidth]{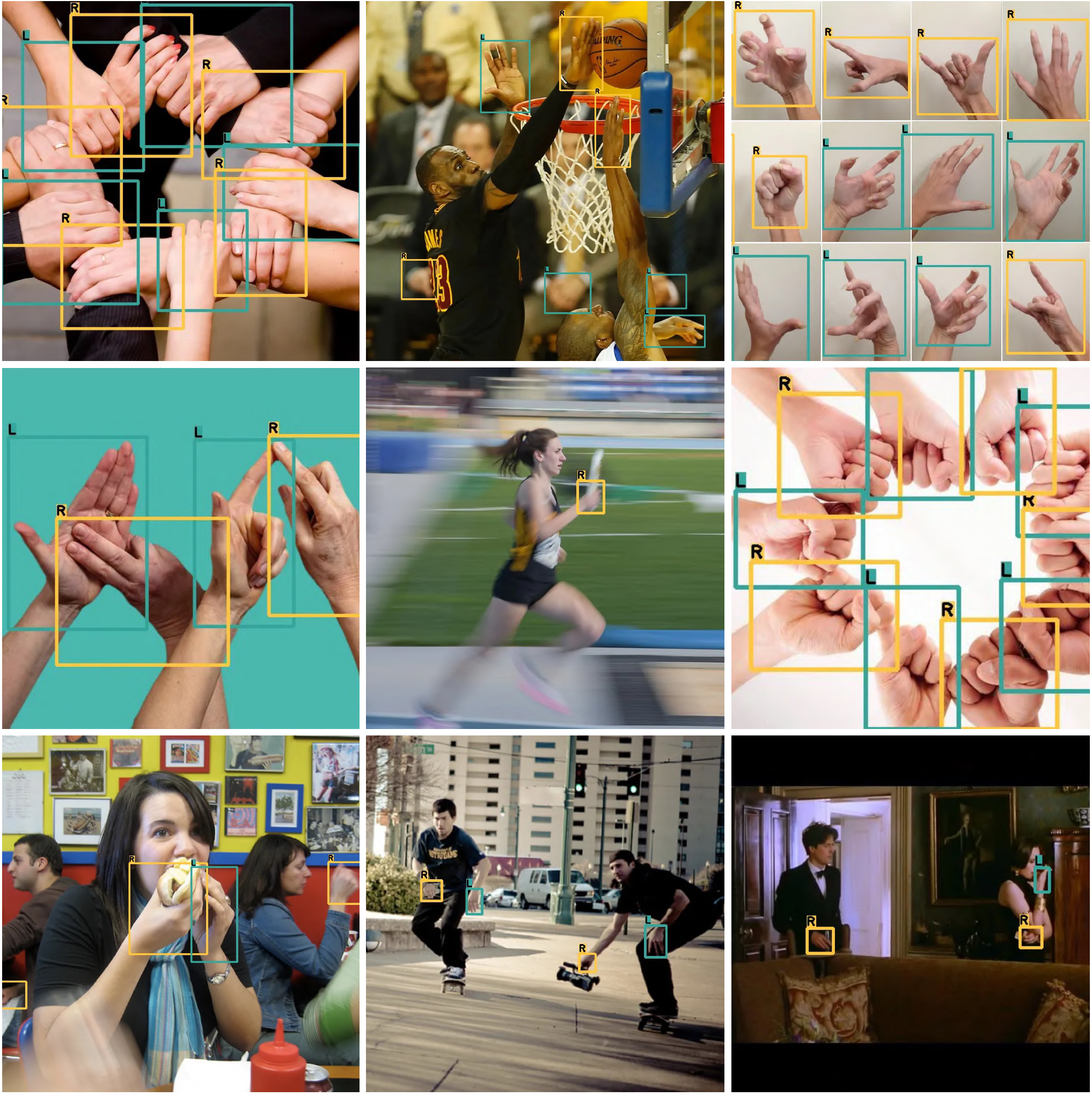}
    \captionof{figure}{\textbf{Qualitative Evaluation} of the proposed hand detection network on in-the-wild images. The proposed model demonstrates robustness across various lighting conditions, resolutions, hand scales, and even in the presence of motion blur.
    \label{fig:detection_qualitative}}
\end{figure}
In \cref{tab:detection}, we evaluate the proposed and the baseline methods on three datasets: the proposed WHIM dataset along with the benchmark Coco-WholeBody~\cite{jin2020whole} and Oxford-Hands~\cite{mittal2011hand} dataset. All experiments were conducted on a NVIDIA RTX 4090 GPU. As can be easily seen, the proposed medium size detector, \textit{Proposed-M}, can run at more than 130 FPS while the small version, \textit{Proposed-S}, can achieve up to 175 FPS,  improving the mAP metric, on average, by 26\% compared to previous state-of-the-art model.
In addition, compared to previous state-of-the-art model, ContactHands~\cite{narasimhaswamy2019contextual}, the proposed detector is $45\times$ faster and has $32\times$ reduced model size which enables the utilization of the proposed detector in mobile applications and heavy pipelines without posing any significant overhead. It is important to note that despite the varying resolutions and hand scales across the three datasets, the proposed model consistently outperforms the baseline methods. This is particularly evident on the COCO-WholeBody dataset~\cite{jin2020whole}, an extension of the COCO dataset that includes full-body images, where the hands are relatively small compared to the overall image size. 
\begin{table}[!t]
  \resizebox{\columnwidth}{!}{
  \centering
  \small
\begin{tabular}{@{}lcc|cc|cc|cc@{}}
& & & \multicolumn{2}{c|}{Coco-Whole} & \multicolumn{2}{c}{Oxford-Hands} & \multicolumn{2}{|c}{WHIM}    \\ 
\toprule
{\footnotesize Method} & {\footnotesize Size (Mb)$\downarrow$  } & {\footnotesize FPS $\uparrow$ } & {\footnotesize AP0.5 $\uparrow$} &  {\footnotesize mAP $\uparrow$} & {\footnotesize AP0.5 $\uparrow$} & {\footnotesize mAP $\uparrow$} &{\footnotesize AP0.5 $\uparrow$} & {\footnotesize mAP$\uparrow$} \\ \midrule

MediaPipe \cite{zhang2020mediapipe} & \cellcolor{tabthird}25 & 25 &       15.43                   &   3.72 & 8.72  & 1.80 & 53.09  & 12.01 \\ 
OpenPose \cite{cao2017realtime}& 141  & \cellcolor{tabthird}29 &       37.05                   &                     9.06   & 20.74  & 4.41 & 76.8  & 34.25 \\ 
ContactHands \cite{narasimhaswamy2020detecting}& 819  & 3& \cellcolor{tabsecond}50.29  & \cellcolor{tabthird}16.67 & \cellcolor{tabthird}70.02  & \cellcolor{tabthird}36.41 & \cellcolor{tabsecond}93.42 & \cellcolor{tabsecond}49.44\\
ViTDet \cite{li2022exploring}& 1400  & 1 & 41.64 &  13.21 & 67.56 & 29.77 & 84.76 & 35.42 \\ \hline
\textbf{Proposed-S}& \cellcolor{tabfirst}\textbf{7}($\times3.5 \downarrow$)  & \cellcolor{tabfirst}\textbf{175}($\times6 \uparrow$) &    \cellcolor{tabthird}46.96  &   \cellcolor{tabsecond}18.56   & \cellcolor{tabsecond}75.21   &  \cellcolor{tabsecond}38.16  &  \cellcolor{tabthird}91.80  &   \cellcolor{tabthird}46.50       \\
\textbf{Proposed-M} & \cellcolor{tabsecond}25 & \cellcolor{tabsecond}138 &  \cellcolor{tabfirst}\textbf{62.48}   & \cellcolor{tabfirst}\textbf{25.97 }  & \cellcolor{tabfirst}\textbf{82.64 } & \cellcolor{tabfirst}\textbf{48.98}  & \cellcolor{tabfirst}\textbf{96.06}  & \cellcolor{tabfirst}\textbf{53.79} \\
\bottomrule
\end{tabular}
}

    \caption{\textbf{Comparison with the state-of-the-art hand detection methods on COCO-Whole~\cite{jin2020whole}, Oxford-Hands~\cite{mittal2011hand} and the proposed WHIM dataset.}  For each method we report the average precision (AP) at IoU=0.5 along with the mean average precision (mAP). We also compare the performance of each method in terms of model size, measured in Mb, and speed, measured in frames per second (FPS).  
    } 
  \label{tab:detection}%
\end{table}%

\noindent{\textbf{Ablation.}}
The efficiency and accuracy of the proposed hand detection method are mainly attributed to the selection of the backbone architecture and the utilization of a large-scale training dataset. We further evaluate the contribution of each component using an ablation study on OxfordHands~\cite{mittal2011hand} and WHIM datasets where we utilized different backbone networks and training datasets. As can be observed from \cref{tab:detection_ablation}, training the detection network with different backbones, apart from achieving similar detection performance, significantly degrades the inference speed of the network. 
The importance of the proposed large-scale in-the-wild WHIM dataset is also validated in \cref{tab:detection_ablation}, where we can observe a significant performance drop when the model was trained with significantly less data, \eg, \textit{Proposed w. 0.25M, Proposed w. 0.5M, Proposed w. 1M}. 
An interesting observation highlighting the versatility of the WHIM dataset is that the proposed model achieves better performance when trained on WHIM compared to the OxfordHands dataset. 
Finally, we evaluate the contribution of the proposed augmentation strategy and the use of landmark regression loss. The augmentation strategy significantly contributes to cross-dataset generalization, achieving 14\% increase on mAP. Similarly, we can observe that incorporating landmark regression loss enhances the detector's precision, leading to more robust detections. 

\addtolength{\tabcolsep}{-5pt}
\begin{table}[!t]
  \resizebox{\columnwidth}{!}{
  \centering
  \small
\begin{tabular}{@{}lcc|cc|cc@{}}
& & & \multicolumn{2}{c}{OxfordHands} & \multicolumn{2}{c}{WHIM}    \\ 
\toprule 
{\footnotesize Method} & {\footnotesize Size (Mb)  $\downarrow$ } &{\footnotesize FPS  $\uparrow$} & {\footnotesize AP0.5 $\uparrow$} &  {\footnotesize mAP $\uparrow$} & {\footnotesize AP0.5 $\uparrow$} &  {\footnotesize mAP $\uparrow$} \\ 
\midrule
Proposed-w. 0.25M & - & - & 49.15 &  26.69 & 75.75 &  38.48\\
Proposed-w. 0.5M & - & - & 58.32 &  35.15 & 83.03 &  42.11\\
Proposed-w. 1M & - & - & 69.21 &  43.04 & 88.37 &  47.92 \\
Proposed-w.OxfordHands & - & - & 68.15 &  40.34 & 70.14 &  35.29\\
\hline
Proposed-w. ResNet50    & \cellcolor{tabsecond}118  & \cellcolor{tabsecond}34  & 74.13 &  47.34 & 95.43 &  51.85\\
Proposed-w. HRNet       & 132 & 30  & \cellcolor{tabfirst}\textbf{84.82} & \cellcolor{tabfirst}\textbf{49.83} & \cellcolor{tabfirst}\textbf{97.23} &  \cellcolor{tabfirst}\textbf{54.12}\\
\hline
Proposed-w/o Augmentations & - & - & 70.76 &  42.17 & 91.13 &  49.93\\
Proposed-w/o Landmark Loss & - & - & 72.57 &  45.96 & 92.43 &  51.44\\
\hline
\textbf{Proposed-M} & \cellcolor{tabfirst}\textbf{25} ($\downarrow 5\times$) & \cellcolor{tabfirst}\textbf{138} ($\uparrow 4\times$) & \cellcolor{tabsecond}{82.64 } & \cellcolor{tabsecond}{48.98}  & \cellcolor{tabsecond}{96.06}  & \cellcolor{tabsecond}{53.79}\\
\bottomrule
\end{tabular}
}  
    \caption{\textbf{Ablation study}: Evaluation of individual components in the proposed detection pipeline on OxfordHands and WHIM datasets. We use $-$ to denote identical network architecture and performance.} 
    \vspace{-0.5cm}
  \label{tab:detection_ablation}%
\end{table}%
\addtolength{\tabcolsep}{5pt}

\subsection{Evaluation of 3D Hand Pose Estimation}
\noindent{\textbf{Training.}} Following \cite{smpler-x,OSX,pavlakos2024reconstructing}, we trained the proposed hand regressor using a combination of datasets to improve robustness to diverse poses, illuminations and occlusions. Particularly, we utilized a set of datasets containing both 2D and 3D annotations namely FreiHAND~\cite{freihand}, HO3D~\cite{hampali2020honnotate},
MTC~\cite{xiang2019monocular}, 
RHD~\cite{zimmermann2017learning}, InterHand2.6M~\cite{Moon_2020_InterHand}, H2O3D~\cite{hampali2020honnotate},
DEX YCB~\cite{chao2021dexycb},
COCO WholeBody~\cite{jin2020whole},
Halpe~\cite{alphapose}
MPII NZSL~\cite{simon2017hand}, BEDLAM~\cite{black2023bedlam},
ARCTIC~\cite{fan2023arctic}, 
Re:InterHand ~\cite{moon2023reinterhand} and 
Hot3D ~\cite{banerjee2024introducing}. In total we utilized 4.2M images, 55\% more than previous state-of-the-art. 

\begin{figure*}[!h]
    \centering
\includegraphics[width=0.93\linewidth]{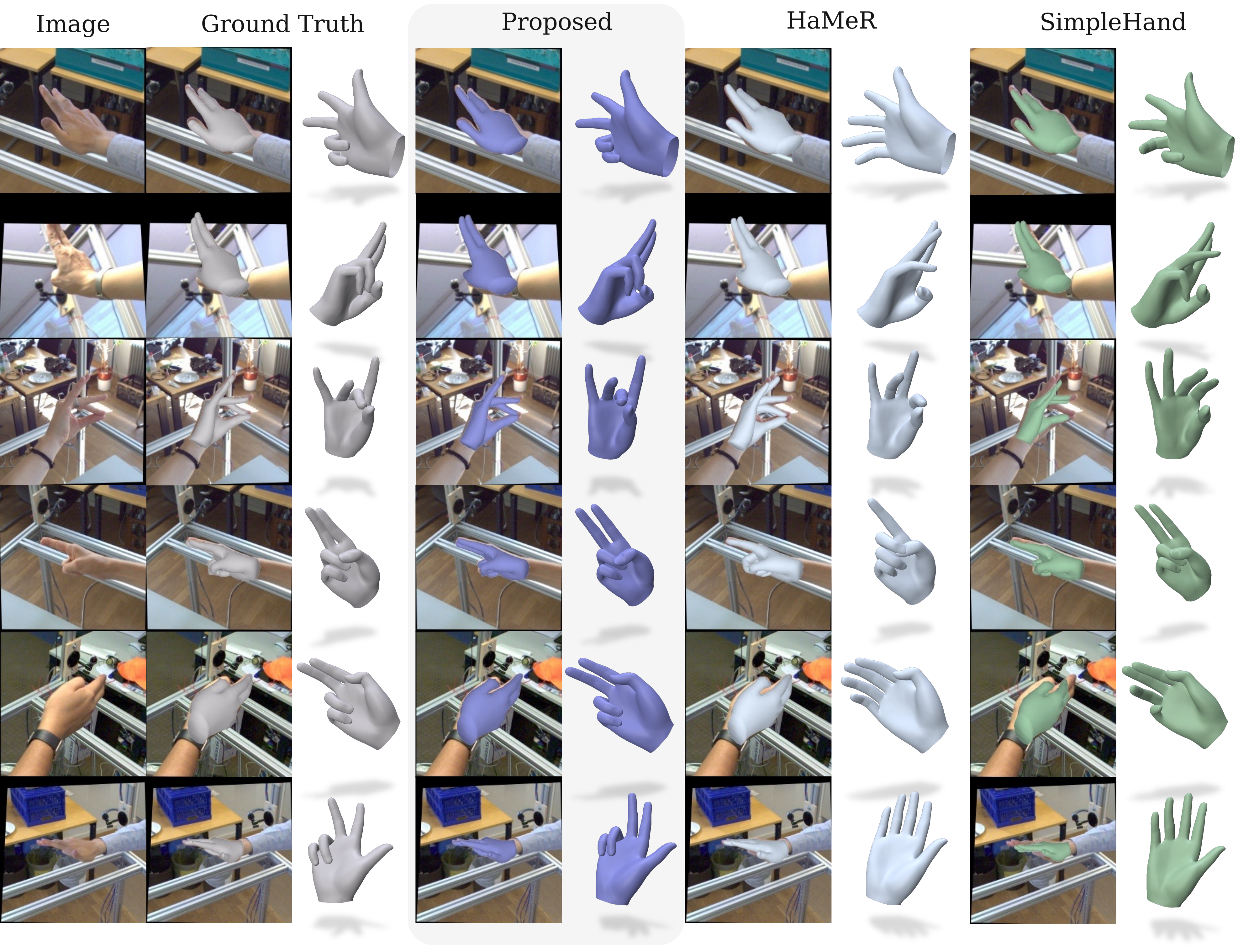}
    \captionof{figure}{\textbf{Qualitative Evaluation} of proposed and the baseline methods on FreiHAND dataset \cite{freihand}. WiLoR demonstrates robustness across challenging poses with heavy occlusions, while maintaining precise image alignment.
    \label{fig:qualitative_freihand}}
\end{figure*}

\noindent{\textbf{Evaluation.}} To compare the proposed method we employ with state-of-the-art methods including METRO~\cite{lin2021end}, 
Mesh Graphormer~\cite{lin2021mesh}, 
AMVUR~\cite{jiang2023probabilistic},
MobRecon~\cite{MobRecon}, HaMeR~\cite{pavlakos2024reconstructing} and SimpleHand~\cite{simpleHand}.
In \cref{tab:freihand} and \cref{tab:ho3d} we report the reconstruction results the popular benchmark FreiHAND~\cite{freihand} and HO3Dv2~\cite{hampali2020honnotate} datasets. Following the common protocol~\cite{freihand}, we measure the reconstruction performance in terms of Procrustes Aligned Mean per Joint and Vertex Error (PA-MPJPE, PA-MPVPE) along with the fraction of poses with less than 5mm and 15mm error (F@5, F@15). Additionally, we report Area Under the Curve for 3D joints and vertices ($\textrm{AUC}_\textrm{J}$, $\textrm{AUC}_\textrm{V}$) for HO3D dataset. 
The proposed method achieves state-of-the-art performance and outperforms previous methods under all metrics on both benchmark datasets, which can be further validated qualitatively in \cref{fig:qualitative_freihand}. 
Leveraging the image-aligned features of the refinement module, WiLoR achieves high fidelity reconstructions even in challenging articulations. 

\addtolength{\tabcolsep}{-5pt}
\begin{table}[!t]
  \resizebox{\columnwidth}{!}{
  \centering
  \small
\begin{tabular}{@{}lcccc@{}}
\toprule
{\footnotesize Method} & {\footnotesize PA-MPJPE $\downarrow$} & {\footnotesize PA-MPVPE $\downarrow$} & {\footnotesize F@5 $\uparrow$} & {\footnotesize F@15 $\uparrow$} \\ \midrule
I2L-MeshNet~\cite{moon2020i2l} & 7.4 & 7.6 & 0.681 & 0.973 \\
Pose2Mesh~\cite{choi2020pose2mesh} & 7.7 & 7.8 & 0.674 & 0.969 \\
I2UV-HandNet~\cite{chen2021i2uv} & 6.7 & 6.9 & 0.707 & 0.977 \\
METRO~\cite{lin2021end} & 6.5 & 6.3 & 0.731 & 0.984 \\
Tang~\etal~\cite{tang2021towards} & 6.7 & 6.7 & 0.724 & 0.981 \\
Mesh Graphormer~\cite{lin2021mesh} & \cellcolor{tabthird}5.9 & 6.0 & 0.764 & 0.986 \\ 
MobRecon~\cite{MobRecon} &  \cellcolor{tabsecond}5.7 & \cellcolor{tabthird}5.8 & \cellcolor{tabthird}0.784 & 0.986 \\
AMVUR~\cite{jiang2023probabilistic} & 6.2 & 6.1 & 0.767 & \cellcolor{tabthird}0.987 \\ 
HaMeR~\cite{pavlakos2024reconstructing} & 6.0 & \cellcolor{tabsecond}5.7 & \cellcolor{tabsecond}0.785 & \cellcolor{tabsecond}0.990  \\ 
\midrule
\textbf{Proposed} & \cellcolor{tabfirst}{\bf 5.5} & \cellcolor{tabfirst} {\bf 5.1} & \cellcolor{tabfirst} {\bf 0.825} &  \cellcolor{tabfirst} {\bf 0.993} \\ 
\midrule
\bottomrule
\end{tabular}
}

    \caption{\textbf{Comparison with the state-of-the-art on the FreiHAND dataset~\cite{freihand}.} We use the standard protocol and
    report metrics for evaluation of 3D joint and 3D mesh accuracy. PA-MPVPE and PA-MPJPE numbers are in mm.}
    \vspace{-0.5cm}
  \label{tab:freihand}%
\end{table}%

\begin{table}[!t]
\resizebox{\columnwidth}{!}{
  \centering
  \scriptsize
\begin{tabular}{@{}lcccccc@{}}
\toprule
Method & $\textrm{AUC}_\textrm{J}$ $\uparrow$ & PA-MPJPE $\downarrow$ & $\textrm{AUC}_\textrm{V}$ $\uparrow$ & PA-MPVPE $\downarrow$ & F@5 $\uparrow$ & F@15 $\uparrow$ \\ \midrule
Liu~\etal~\cite{liu2021semi} & 0.803 & 9.9 &0.810&9.5&0.528&0.956\\
HandOccNet~\cite{park2022handoccnet} & 0.819 & 9.1 &0.819&8.8&0.564&0.963\\
I2UV-HandNet~\cite{chen2021i2uv} & 0.804 & 9.9 &0.799&10.1&0.500&0.943\\
Hampali~\etal~\cite{hampali2020honnotate} & 0.788 & 10.7 &0.790&10.6&0.506&0.942\\
Hasson~\etal~\cite{hasson2019learning} & 0.780 & 11.0 &0.777&11.2&0.464&0.939\\
ArtiBoost~\cite{yang2022artiboost} & 0.773 & 11.4 &0.782&10.9&0.488&0.944\\
Pose2Mesh~\cite{choi2020pose2mesh} & 0.754 & 12.5 &0.749&12.7&0.441&0.909\\
I2L-MeshNet~\cite{moon2020i2l} & 0.775 & 11.2 &0.722&13.9&0.409&0.932\\
METRO~\cite{lin2021end} & 0.792 & 10.4 & 0.779 & 11.1 & 0.484 & 0.946\\
MobRecon\cite{MobRecon}& -&9.2&-& 9.4& 0.538& 0.957\\
Keypoint Trans~\cite{hampali2022keypoint} & 0.786 & 10.8 &-&-&-&-\\
AMVUR~\cite{jiang2023probabilistic} & \cellcolor{tabthird} 0.835 & \cellcolor{tabthird} 8.3 & \cellcolor{tabthird} 0.836 & \cellcolor{tabthird} 8.2 & \cellcolor{tabthird} 0.608 & \cellcolor{tabthird} 0.965 \\ 
HaMeR & \cellcolor{tabsecond} {0.846} & \cellcolor{tabsecond} {7.7} & \cellcolor{tabsecond} { 0.841} & \cellcolor{tabsecond} { 7.9} & \cellcolor{tabsecond} {0.635} & \cellcolor{tabsecond} {0.980} \\ 
\midrule
\textbf{Proposed} & \cellcolor{tabfirst} {\bf 0.851} & \cellcolor{tabfirst} {\bf 7.5} & \cellcolor{tabfirst} {\bf 0.846} & \cellcolor{tabfirst} {\bf 7.7} & \cellcolor{tabfirst} {\bf 0.646} & \cellcolor{tabfirst} {\bf 0.983 } \\ 
\bottomrule
\end{tabular}
}
    \caption{\textbf{Comparison with the state-of-the-art on the HO3D dataset~\cite{hampali2020honnotate}.} We use the HO3Dv2 protocol and report metrics that evaluate accuracy of the estimated
    3D joints and 3D mesh. PA-MPVPE and PA-MPJPE numbers are in mm.}
  \label{tab:ho3d}%
\end{table}%

\begin{table}[!t]
  \resizebox{\columnwidth}{!}{
  \centering
  \scriptsize
\begin{tabular}{@{}lcccc@{}}
\toprule
{\footnotesize Method} & {\footnotesize PA-MPJPE $\downarrow$} & {\footnotesize PA-MPVPE $\downarrow$} & {\footnotesize F@5 $\uparrow$} & {\footnotesize F@15 $\uparrow$} \\ 
\midrule
Proposed w. FastViT  & 6.5 & 6.3 & 0.741 & 0.967 \\
Proposed w/o ViTPose &  5.9 & 5.7 & 0.795 &  0.989 \\
\midrule
Proposed w. Single-Scale & 6.0 & 5.9 & 0.793 &  0.991 \\ 
Proposed w/o Refinement & 6.1  & 5.8 & 0.795 &  0.991 \\
\midrule
Proposed w. FreiHAND \cite{freihand} & 6.1 & 5.8 & 0.793 &  0.990 \\
Proposed w. Datasets \cite{pavlakos2024reconstructing}  & 5.9 & 5.7 & 0.805 &  0.992 \\
\midrule
\textbf{Proposed Full} &  \cellcolor{tabfirst}{\bf 5.5} & \cellcolor{tabfirst} {\bf 5.1} & \cellcolor{tabfirst} {\bf 0.825} &  \cellcolor{tabfirst} {\bf 0.993} \\ 
\bottomrule
\end{tabular}
}

\caption{\textbf{Ablation study on the FreiHAND dataset~\cite{freihand}.} We report ablations on the backbone and the training data used along with the novel refinement module.}
\vspace{-0.5cm}
\label{tab:reconstruction_ablation}%
\end{table}%

\noindent{\textbf{Ablation.}} To further investigate the contributions of each component of the proposed method we conducted an ablation study. In \cref{tab:reconstruction_ablation}, we assess the contribution of the backbone architecture, the training datasets used along with the refinement module. As can be observed, swapping the ViT backbone with the recent efficient FastViT~\cite{vasu2023fastvit} architecture \textit{(Proposed w. FastViT)} results in significant degradation of performance despite the runtime efficiency. Similarly, training the backbone from scratch without using the pre-trained weights of ViTPose~\cite{xu2022vitpose} \textit{(Proposed w/o ViTPose)} also results in a performance drop. To evaluate the effect of the proposed refinement module we trained a model that directly regresses the MANO and camera parameters from the ViT output tokens without using any refinement module \textit{(Proposed w/o Refinement)}. Additionally, we trained a model with a single-scale refinement module that samples features from a single feature map \textit{(Proposed w. Single-Scale)}. Both architectural choices deteriorate the reconstruction performance of the proposed model which highlights the effect of the proposed multi-scale refinement module. Finally, we examine the effect of the large-scale training set by training two derivatives of the proposed model using only the FreiHAND dataset~\cite{freihand} \textit{(Proposed w. FreiHAND~\cite{freihand})}, similar to~\cite{lin2021end,lin2021mesh,simpleHand,MobRecon} and a model trained on the datasets used in~\cite{pavlakos2024reconstructing} \textit{(Proposed w. Datasets~\cite{pavlakos2024reconstructing})}. 

\subsection{Evaluation of Dynamic Reconstruction}
A key challenge for 3D pose estimation methods is to achieve stable and robust 4D reconstructions without being trained using a dynamic setting~\cite{shin2024wham}. Traditionally, methods for 3D pose estimation from single image suffer from low temporal coherence and jittering effects across frames, setting a huge burden on their generalization to real-world video reconstruction. To effectively evaluate the temporal coherence of the proposed method, we reconstructed frame-wise a 4D sequence and measure the jittering between frames. In particular, we calculate the mean per frame Euclidean distance of the 3D vertices (MPFVE) and joints (MPFJE) between consecutive frames. Additionally, similar to~\cite{shin2024wham}, we measure the jerk (Jitter) of the 3D hand joints motion along with the global Root Translation Error (RTE) that measures the displacement of the wrist across frames. In \cref{tab:jitter}, we report the reconstruction results for the best performing methods on HO3D~\cite{hampali2020honnotate} dataset. WiLoR outperforms baseline methods in temporal coherence without relying on any temporal module based on the robust stability of the detections. We refer the reader to the supplementary material for qualitative video results.

\begin{table}[!t]
  \resizebox{\columnwidth}{!}{
  \centering
  \scriptsize
\begin{tabular}{@{}lcccc@{}}
\toprule
{\footnotesize Method} & {\footnotesize MPFVE $(\times 100) \downarrow$} & {\footnotesize MPFJE $(\times 100) \downarrow$} & {\footnotesize Jitter $\downarrow$} &  {\footnotesize RTE $\downarrow$}\\ 
\midrule
MeshGraphormer \cite{lin2021mesh}  & 21.86 &  4.99 & 41.16  & 7.92 \\
MobRecon \cite{MobRecon}  & 22.18 &  6.09 & 40.25  & 8.03 \\
SimpleHand \cite{simpleHand}  &\cellcolor{tabthird}  19.72 &  \cellcolor{tabthird} 5.12 & \cellcolor{tabthird} 38.53  & \cellcolor{tabthird} 6.04 \\
HaMeR \cite{pavlakos2024reconstructing}  & \cellcolor{tabsecond} 10.60 &  \cellcolor{tabsecond} \cellcolor{tabsecond} 1.768 & \cellcolor{tabsecond} 20.43 &\cellcolor{tabsecond} 2.92 \\
\midrule
\textbf{Proposed} & \cellcolor{tabfirst} {\bf 4.43} & \cellcolor{tabfirst}  {\bf 0.762} &  \cellcolor{tabfirst}   {\bf 5.92} &  \cellcolor{tabfirst}  {\bf 0.07} \\
\bottomrule
\end{tabular}
}
\caption{\textbf{Reconstruction of dynamic 3D Hands.} We evaluate the temporal coherence and the jittering of the reconstruction for the proposed and the baseline methods on the HO3D dataset. 
}
\label{tab:jitter}%
\end{table}%

\section{Conclusion}
In this work, we present WiLoR, the first full-stack hand detection and 3D pose estimation framework. 
Using a large-scale in-the-wild dataset we train a light-weight yet highly accurate hand detector model that can robustly detect hands under different occlusions and illuminations at over 130 FPS. 
Additionally, we propose a high fidelity 3D hand pose estimation model built on top of our novel refinement module, that overcomes the limitations of previous methods and mitigates the alignment issues of previous methods. Under a series of experiments, we showcase that WiLoR outperforms previous state-of-the-art methods on two benchmark datasets and show robust performance on challenging cases. 
WiLoR establishes a comprehensive solution for multi-hand detection, localization and 3D reconstruction. 

\noindent\textbf{Acknowledgements} S. Zafeiriou was supported by Turing AI Fellowship (EP/Z534699/1) and GNOMON (EP/X011364). R.A. Potamias was supported by EPSRC Project GNOMON (EP/X011364).
{
    \small
    \bibliographystyle{ieeenat_fullname}
    \bibliography{main}
}

\clearpage
\clearpage
\setcounter{page}{1}
\maketitlesupplementary

\section{Implementation Details}
In this section we report the training details of the hand detection and the hand pose estimation models. 
\subsection{Hand Detection and Localization}
To train the detector model we use WHIM dataset that comprises of over 2M in the wild images from daily activities. 
We train WiLoR detector with Adam optimizer for 200 epochs with early stopping if there is no loss decrease for over 30 epochs. 
Initiate the training with a learning rate of 0.01 and linearly decrease to $1e-6$ for the last 30 epochs of the training. 
We trained the model for three weeks using two NVIDIA RTX 4090 and a batch size of 256. 
To weight the different losses we set $\lambda_0 = 0.5 $ for the classification loss, $\lambda_1 = 1.5 $ for the distribution focal length loss, $\lambda_2 = 15$ for the bounding box loss, $\lambda_3 = 10$ for the keypoints loss. 
We use random mosaic augmentations with probability 0.7, random rotations between $[-60^o, 60^o]$ and random image scaling between [0.5, 1].

\subsection{Hand Pose Estimation}
We build our hand pose estimation method on top of a ViT-Large backbone with pre-trained weights from ViTPose~\cite{xu2022vitpose}, with a hidden dimension of 1280.
Apart from the image patches, we use three additional learnable tokens that correspond to hand pose, shape and camera translation and scale. 
We initialize the tokens with the mean pose, shape and camera parameters from the training set. 
Using a set of fully-connected layers, we map the output tokens to 96 MANO pose parameters (15 joint rotations + 1 global orientation represented in 6d rotation format ~\cite{zhou2019continuity}), 10 MANO shape parameters and a 3D camera translation. 
We then reshape the output image tokens to a $16\times12$ image form, and perform two sets of upsamplings using deconvolutions.
At each upsampling step we reduce the feature by 2 times.
Using the initially estimated camera parameters we project the rough MANO estimation to the feature maps and sample a set of multi-scale per-vertex image-aligned features. 
The concatenated set of features is then aggregated and regressed from a set of fully-connected layers that predict the pose, shape and camera residuals. 
We train the model for 1000 epochs using Adam optimizer with an initial learning rate of 1e-5 and a weight decay of 1e-4. 
Similar to the hand detector, we apply random scaling, rotations and color jitter during training. 
Similar to ~\cite{pavlakos2024reconstructing}, to balance the losses we set $\lambda_{3D} = 0.05$, $\lambda_{2D} = 0.01$, $\lambda_{pose} = 0.001$, $\lambda_{shape} = 0.0005$ and $\lambda_{adv} = 0.0005$. 

\section{Comparison with existing datasets}
In contrast to 3D hand pose estimation methods that utilizes images of tightly cropped hands, to train a powerful hand detector network, it is required to create a dataset that contains images with multiple hands under different occlusions, views, illuminations and skin tones. Bellow we compare WHIM with such available datasets. WHIM is 100$\times$ larger than previous in-the-wild multi-hand datasets. 
\begin{table}[!h]
  \centering
  \resizebox{\linewidth}{!}{
  \small
\begin{tabular}{@{}l|ccccccc@{}}
\toprule 
{\footnotesize Dataset} & {\footnotesize \#Img}& {Annotations}&{Egocentric} &{Third-Person} & {Objects} & {Real} &{3D} \\ 
\midrule
OxfordHands  & 13K & Manual& \ding{56} & \ding{52} &  \ding{56} &\ding{52} & \ding{56} \\
MPI-HP  & 25K& Manual & \ding{56} & \ding{52} &  \ding{56} &\ding{52} & \ding{56} \\
Coco-Whole  & 200K & Manual & \ding{56} & \ding{52} &  \ding{56} &\ding{52} & \ding{56} \\
BEDLAM   & 380K & GT & \ding{56} & \ding{52} &  \ding{56} &\ding{56} & \ding{52} \\
AGORA   & 18K & GT& \ding{56} & \ding{52} &  \ding{56} &\ding{56} & \ding{52} \\
ContactHands  & 21K & Manual& \ding{56} & \ding{52} &  \ding{52} &\ding{52} & \ding{56} \\
CocoHands  & 25K & Manual & \ding{56} & \ding{52} &  \ding{52} &\ding{52} & \ding{56}\\
BodyHands  & 20K & Manual & \ding{56} & \ding{52} &  \ding{56} &\ding{52} & \ding{56}\\
\bottomrule
\textbf{WHIM}& \textbf{2M} & Auto & \ding{52} & \ding{52} & \ding{52} &\ding{52} & \ding{52} \\
\bottomrule 
\end{tabular}
}
\caption{\textbf{Comparison with existing hand datasets.} WHIM is 100$\times$ larger than previous multi-hand dataset.s}
\end{table}    

\section{Limitations}
\begin{figure*}[!h]
    \centering
\includegraphics[width=\linewidth]{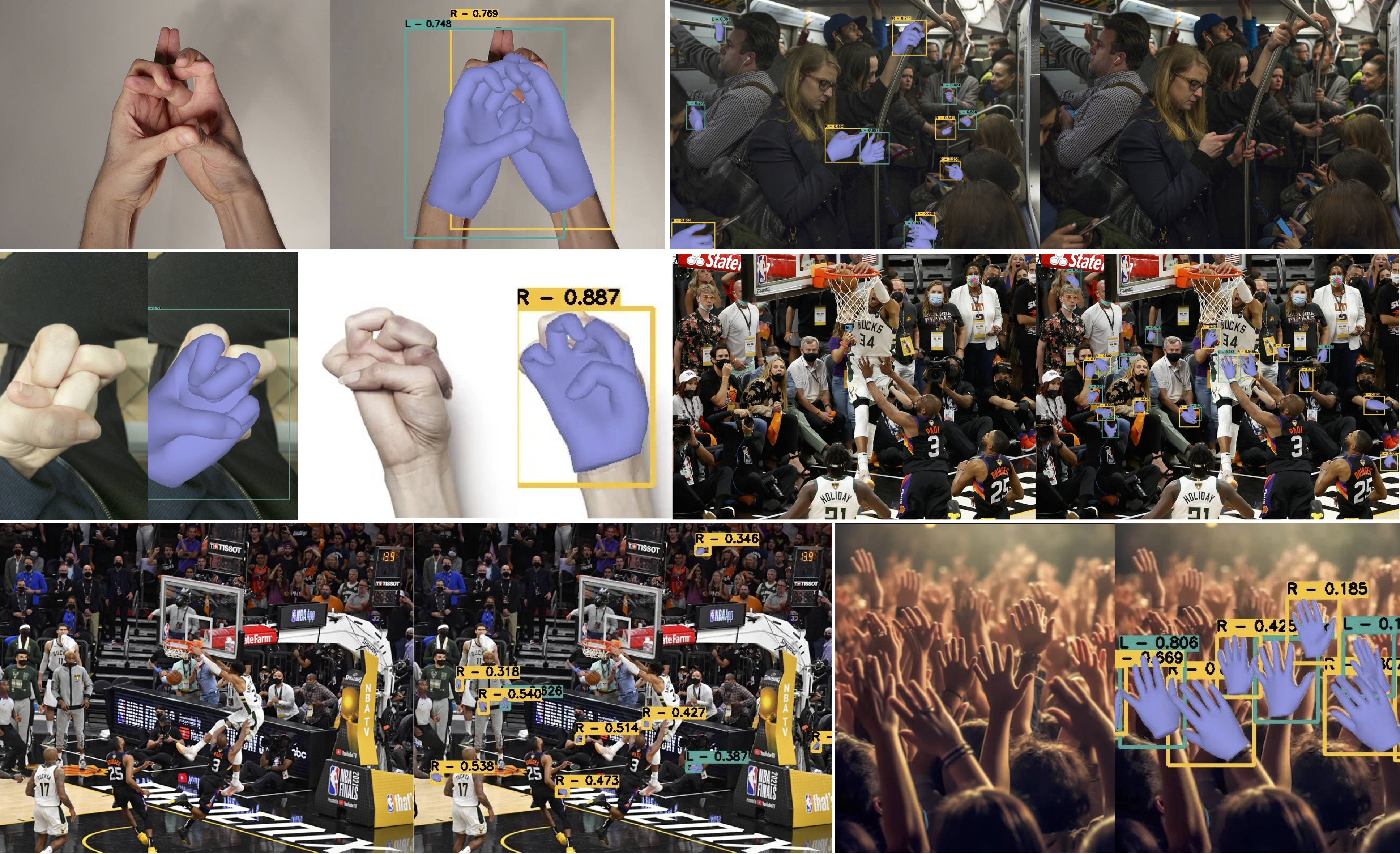}
    \captionof{figure}{\textbf{Failure Cases}. WiLoR can still fail to reconstruct complex finger poses or detect small hands in crowded environment. 
    \label{fig:failure}}
\end{figure*}

Although achieving state-of-the-art performance on both 3D hand pose estimation and hand detection tasks, WiLoR still fails to recover challenging cases. Despite being trained on a large-scale dataset, the data distribution is still limited to `common' hand poses and appearances, failing to generalize to samples far from the trained distribution. As can be seen in \cref{fig:failure} WiLoR can fail under extreme finger poses and can also fail to detect hands in crowded environments. 
Creating a synthetic dataset with diverse hand poses and photorealistic hands could help mitigate these issues ~\cite{Potamias_2023_CVPR}.
Additionally, since WiLoR employs a bottom-up reconstruction strategy, interactions and contacts between hands may not be adequately captured in 3D space. In scenarios where accurate hand contact estimation is crucial ~\cite{Baltatzis_2024_CVPR,zuo2024signs}, incorporating additional interaction constraints \cite{xie2024ms} may be necessary.
Finally, WiLoR estimates 3D hand poses in camera space, which may lead to inaccurate assumptions about the overall 3D scene. Adapting WiLoR with a 3D metric foundational model \cite{yin2023metric3d,zhang2025hawor} could enable more accurate 3D reconstruction in world space.

\section{Training Datasets}
To train our hand pose estimation module we use a combination of datasets to enforce the generalization of the model. 
In particular, we use 14 datasets with both 2D and 3D annotations, from three major categories: controlled environment hand images, hand-object interaction, in-the-wild and synthetic datasets, resulting in 4.2M images total: 
\begin{itemize}
    \item FreiHAND~\cite{freihand} is a common 3D hand pose estimation dataset composed of 132K images of indoor, outdoor, and synthetic scenes. It provides both 3D hand and 2D keypoint annotations. 
    \item MTC~\cite{xiang2019monocular} is a subset of Panoptic Studio Dataset \cite{Joo_2017_TPAMI} that contains 360K multi-view images in a studio environment. The dataset provides both 3D hand and 2D keypoint annotations. 
    \item InterHand2.6M~\cite{Moon_2020_InterHand} is a large scale environment from light stage environment that contains hand articulations from 27 different subjects and 80 different cameras. The dataset provides both 3D hand and 2D keypoint annotations. 
\end{itemize}
To increase the generalization of WiLoR under severe occlusions we include several datasets where hands interact with objects. 
\begin{itemize}
    \item HO3D~\cite{hampali2020honnotate} provides a hand-object dataset with over than 120K images from multi-view cameras of hands interacting with objects. It is used as one of the main benchmarks for hand and object reconstruction. Images were captured in an lab environment setting. It provides both 3D hand and 2D keypoint annotations. 
    \item H2O3D~\cite{hampali2020honnotate} contains over 60K images from five multi-view cameras of hands interacting with objects. In contrast to HO3D, each subject is interacting with the object using two hands which increases the occlusions of the hands. It provides both 3D hand and 2D keypoint annotations.
    \item DEX YCB~\cite{chao2021dexycb} similar to HO3D, DEX YCB is a benchmark dataset for hand object reconstruction. It contains over than 500K multi-view images from 10 objects grasping objects. The dataset provides both 3D hand and 2D keypoint annotations.
    \item ARCTIC~\cite{fan2023arctic} is a large scale dataset of bimanual hand-object manipulations containing over than 400K images from both egocentric and third-person views. It contains both single and dual hand manipulations along with accurate 3D and 2D hand annotations. 
    \item Hot3D ~\cite{banerjee2024introducing} is a recent egocentric dataset from daily activities that includes a high degree of occlusions and can significantly enhance the performance of WiLoR in egocentric scenarios. It contains both 3D hand and 2D keypoint annotations. 
\end{itemize}
Additionally, we include four synthetic datasets that provide accurate ground truth 2D and 3D annotations: 
\begin{itemize}
    \item RHD~\cite{zimmermann2017learning} is amongst the first synthetic datasets of hands rendered under different illumination patterns. The dataset is composed of 62K images with accurate 3D and 2D annotations.  
    \item Re:InterHand ~\cite{moon2023reinterhand} is synthetic dataset that extends InterHand2.6M by rendering hands under different illuminations and environments to bridge the gap between studio setup and in-the-wild images. The dataset provides both 3D hand and 2D keypoint annotations.
    \item BEDLAM~\cite{black2023bedlam} is a large scale full body synthetic dataset, that has proven extremely effective in whole body reconstruction tasks~\cite{smpler-x}. We use BEDLAM and randomly crop regions around the human hands to augment the training data. We use over 500K image crops. The dataset provides both 3D hand and 2D keypoint annotations.
\end{itemize}
Finally, we include in-the-wild datasets that contain only 2D information, but can effectively boost the generalization performance of WiLoR in in the while scenarios. 
\begin{itemize}  
    \item COCO WholeBody~\cite{jin2020whole} is a subset of COCO dataset and one of the main benchmarks for body pose estimation. It contains over than 100K in the wild images with humans participating in different activities. It provides 2D hand keypoint annotations. 
    \item Halpe~\cite{alphapose} is a full body in-the-wild dataset, composed of over than 40K images will 2D keypoint annotations. The dataset contains 21 keypoints for each hand. 
    \item MPII NZSL~\cite{simon2017hand} is a common benchmark for human body pose estimation, containing a set of in-the-wild, synthetic and lab environment images. The dataset contain 15K images with 2D keypoint hand annotations.
\end{itemize}

\section{Temporal Coherence}
In the \href{https://rolpotamias.github.io/WiLoR/}{project page} we provide several videos that demonstrate the 
temporal coherence of WiLoR in challenging scenarios such as kneading dough or playing guitar. 
Despite being trained on single images, WiLoR can provide smooth reconstructions given its stable and robust detections.



\end{document}